\documentclass[letterpaper, 10 pt, conference]{ieeeconf}  %

\IEEEoverridecommandlockouts                              %

\usepackage{graphics} %
\usepackage{epsfig} %
\usepackage{times} %
\usepackage{amsmath} %
\usepackage{amssymb}  %

\usepackage{graphicx}
\usepackage{subcaption}

\usepackage{booktabs}

\usepackage{pifont}

\usepackage{xcolor}

\usepackage[export]{adjustbox}%
\usepackage{float}

\usepackage{stfloats}

\title{\LARGE \bf
Unsupervised Skill Discovery for Robotic Manipulation through Automatic Task Generation
}

\author{
Paul Jansonnie$^{12}$, Bingbing Wu$^{1}$, Julien Perez$^{3\dag}$ and Jan Peters$^{2456}$%
\thanks{$^{1}$NAVER LABS Europe, 6 chemin de Maupertuis, Meylan, 38240, France. 
email: {\tt\small firstname.lastname@naverlabs.com}}
\thanks{$^{2}$Department of Computer Science, TU Darmstadt, Germany. email: {\tt\small firstname.lastname@tu-darmstadt.de}}%
\thanks{$^{3}$EPITA Research Laboratory (LRE), FR-94276 Le Kremlin-Bicêtre, France. {\tt\small firstname.lastname@epita.fr}}
\thanks{$^{4}$German Research Center for AI (DFKI), Systems AI for Robot Learning}
\thanks{$^{5}$Center for Cognitive Science, TU Darmstadt, Germany}
\thanks{$^{6}$Hessian Center for Artificial Intelligence (Hessian.ai), Germany}
\thanks{$^{\dag}$Work done at Naver Labs Europe}
}

\begin{document}

\makeatletter
\let\@oldmaketitle\@maketitle%
\renewcommand{\@maketitle}{\@oldmaketitle%
    \centering
    \vspace*{1mm}
    \includegraphics[width=0.85\textwidth]{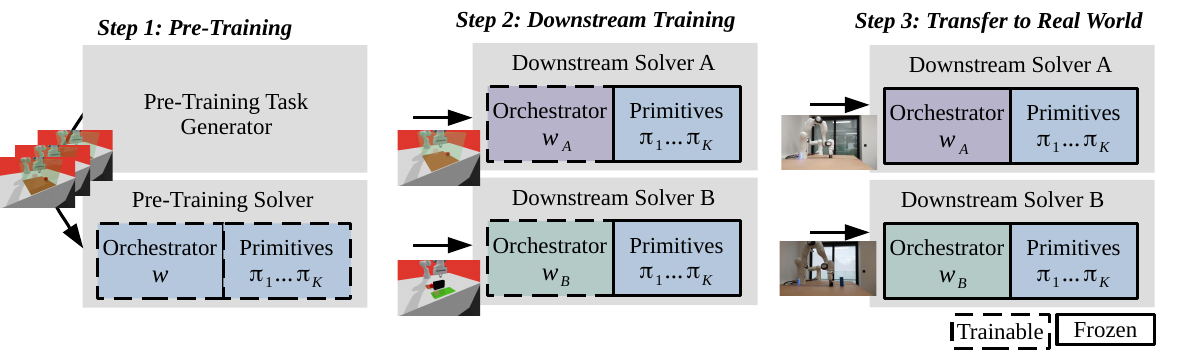} %
    \captionof{figure}{\textbf{Method Overview}:
    A task \textit{solver} is pre-trained in simulation to solve tasks that are autonomously proposed by a task \textit{generator} (left).
    The \textit{solver} discovers diverse behaviors and embeds them in its primitives.
    On each downstream task, an orchestrator is trained to reuse the pre-trained primitives (middle).
    Downstream agents are transferred to a real robotic platform (right).
    }
    \label{fig:method}
    \vspace*{-3mm}
}
\makeatother
\maketitle
\IEEEpeerreviewmaketitle
\setcounter{figure}{1}

\maketitle
\thispagestyle{empty}
\pagestyle{empty}

\begin{abstract}
    Learning skills that interact with objects is of major importance for robotic manipulation.
    These skills can indeed serve as an efficient prior for solving various manipulation tasks.
    We propose a novel Skill Learning approach that discovers composable behaviors by solving a large and diverse number of autonomously generated tasks.
    Our method learns skills allowing the robot to consistently and robustly interact with objects in its environment.
    The discovered behaviors are embedded in primitives which can be composed with Hierarchical Reinforcement Learning to solve unseen manipulation tasks.
    In particular, we leverage Asymmetric Self-Play to discover behaviors and Multiplicative Compositional Policies to embed them.
    We compare our method to Skill Learning baselines and find that our skills are more interactive.
    Furthermore, the learned skills can be used to solve a set of unseen manipulation tasks, in simulation as well as on a real robotic platform.
\end{abstract}

\section{Introduction}

Robotic manipulation is notorious for being a challenging problem for Reinforcement Learning (RL), especially for Goal-Conditioned RL (GCRL).
Manipulation tasks are indeed often associated with \textit{sparse} rewards~\cite{riedmiller2018sacX,vecerik2018robotics}, which dramatically increases the complexity of learning a successful policy.
One classic example is the task of moving an object from an initial position to a target position, also known as a pick-and-place task.
This task involves complex contact reasoning, fine control of the robotic manipulator, and possibly safety constraints.
Learning to solve such a task usually requires reward shaping~\cite{popov2017lego,yu2020meta}, hindsight relabeling~\cite{andrychowicz2017hindsight} or human demonstrations~\cite{nair2018demo}.
An alternative is the use of pre-trained behaviors, also known as skills or primitives.
To cope with the complexity of training manipulation policies from scratch, Hierarchical Reinforcement Learning (HRL) proposes orchestrating pre-trained behaviors.
Reusing a repertoire of pre-trained skills aims at increasing the probability of success, e.g., by making interactions with objects more frequent and consistent.
In HRL, an orchestrator policy is trained to reuse pre-trained skills, which allows for maintaining a usable reward signal from the environment.

Skill Discovery, or Skill Learning, is a family of methods that focus on autonomously discovering behaviors such that they can then be reused, e.g., with HRL.
The most popular approaches rely on task-agnostic intrinsic rewards based on the mutual information between a skill descriptor and a transformation of the state of the environment~\cite{eysenbach2019_diayn,Sharma2020DynamicsAwareUS, achiam2018_valor, gregor2016_vic, warde2019_discern,pong2019_skewfit,park2021lipschitz,pmlr-v202-park23h,laskin2022unsupervised,campos2020explore}.
While these methods usually produce reusable behaviors in locomotion settings, they often fail to produce useful skills for manipulation, i.e., skills that yield interactions between the robot and objects in the scene. %
This property limits their reusability for learning downstream manipulation tasks with HRL.
Furthermore, the discovered skills must not only yield interactions with objects, but these interactions must be diverse for reuse in a large set of downstream tasks. %

This lack of meaningful interactions is the main struggle in Skill Learning for manipulation and is a direct consequence of the sparsity of the intrinsic reward based on mutual information.

In this work, we propose a Skill Learning method for robotic manipulation that discovers behaviors enabling the robot to purposely interact with objects in the environment.
As depicted in Figure \ref{fig:method}, our approach relies on pre-training skills to solve automatically generated tasks, which leads to the discovery of diverse and complex behaviors that can be composed on downstream tasks with HRL. 
We propose to leverage Asymmetric Self-Play (ASP) \cite{sukhbaatar2017intrinsic,sukhbaatar2018learning, OpenAI2021AsymmetricSF} to generate a wide diversity of increasingly difficult tasks and pre-train a Multiplicative Compositional Policy (MCP) \cite{Peng2019MCPLC} to solve the generated tasks.
The discovered behaviors are embedded in the primitives of the policy, which are later reused to solve downstream tasks, that could otherwise not be solved neither with skills learned with current Skill Discovery methods nor with policies trained without priors.
This curriculum induced by ASP enables the skill repertoire to efficiently capture increasingly diverse and complex behaviors in a self-supervised fashion, without the need for complex reward-shaping.
We evaluate the learned skills on a set of unseen downstream manipulation tasks, which are variants of pick-and-place tasks with modifications of the dynamics, such as the introduction of an obstacle in the scene.
We show that our skills achieve higher performance than skills from other approaches on most downstream tasks when orchestrated with HRL.
Our findings demonstrate that learning an orchestrator on top of our pre-trained skills is comparable to or better than ASP and MCP used in isolation on most downstream tasks. %
Finally, we provide evidence that our skills outperform Hindsight Experience Replay \cite{andrychowicz2017hindsight} on several tasks. %
Beyond simulation, we confirm that our approach transfers to a real-world manipulation platform. %

In summary, our contributions are the following: 
(1) we propose a novel approach for reusable Skill Learning based on Asymmetric Self-Play and Multiplicative Compositional Policies,
(2) we perform a qualitative evaluation of the resulting skills,
(3) we evaluate our approach against state-of-the-art Skill Learning approaches on downstream tasks.

\section{Related Work}
\label{sec:rel}

\paragraph{Skill Learning}

A large collection of work studies the problem of learning reusable skills. %
Latent variable methods~\cite{eysenbach2019_diayn,Sharma2020DynamicsAwareUS, achiam2018_valor, gregor2016_vic, warde2019_discern,pong2019_skewfit,park2021lipschitz,pmlr-v202-park23h,laskin2022unsupervised,campos2020explore} learn skills by maximizing the mutual information between a transformation of the environment state $s$ and a latent skill descriptor $z$.
Option architectures~\cite{stolle2002learning, bacon2017option, DBLP:journals/corr/abs-2106-13105} learn to decompose tasks into a set of sub-goals, or options.
Multiplicative Compositional Policies \cite{Peng2019MCPLC} learn to decompose behaviors in distinct Gaussian distributions that can be re-composed to yield more complicated behaviors. %
Our approach is similar to latent variable methods as we learn skills in an unsupervised manner.
However, our training objective is not based on mutual information but rather on the usability of our skills in a manipulation scenario.
Furthermore, contrary to latent variable methods, our approach explicitly learns skills that are composable to solve manipulation tasks.
Our method is most similar to the latter family of approaches as we also rely on an MCP.
However, our work differs from the pre-training objective of the MCP.
While \cite{Peng2019MCPLC} pre-trains primitives from cherry-picked demonstrations and \cite{cho2022unsupervised} learns them using an objective based on mutual information, we pre-train our primitives to maximize their usability on automatically generated tasks.
\paragraph{Pre-Training and Automatic Task Generation}
Many works show the difficulty of learning complex tasks with RL and propose automated curricula~\cite{andrychowicz2017hindsight,florensa2017reverse,salimans2018backward,matiisen2019TSCL,zhang2020automatic,portelas2020teacher} or auxiliary exploration objectives~\cite{oudeyer2007lp,baranes2013motivation,pathak2017curiosity,burda2018exploration,ecoffet2019GoExplore,Ecoffet2020FirstRT} to learn \textit{predefined tasks}.
When training goal-conditioned policies, relabeling or reversing trajectories~\cite{andrychowicz2017hindsight,florensa2017reverse,salimans2018backward} or imitating successful demonstrations~\cite{oh2018SIL,ecoffet2019GoExplore,Ecoffet2020FirstRT} naturally reduces the task complexity.
Open-ended methods~\cite{wang2019poet,wang2020enhanced,du2022takes,dennis2020emergent,parkerholder2022evolving,fang2021discovering,fang2022active,jiang2021replay} automatically generate increasingly complicated tasks to train an agent that discovers behaviors in an environment.
Among these methods is Asymmetric Self-Play ~\cite{sukhbaatar2017intrinsic,sukhbaatar2018learning,OpenAI2021AsymmetricSF}, which was shown to produce diverse tasks and solutions in the context of robotic manipulation. 

A limitation of~\cite{OpenAI2021AsymmetricSF} is that the trained agent uses a monolithic policy, which reduces the reusability of the learned behaviors in new scenarios.
Our approach is most similar to \cite{OpenAI2021AsymmetricSF} but differs in the fact we only use ASP as a pre-training of skills that can then be reused with HRL on novel tasks.

\section{Preliminaries}
\label{sec:pre}
\subsection{Reinforcement Learning and Goal-Conditioning}

We consider an environment with a state space $\mathcal{S}$, an action space $\mathcal{A}$, a state transition probability $p(s_{t+1} | s_t, a_t)$, where $s_t \in \mathcal{S}$, $s_{t+1} \in \mathcal{S}$, $a_t \in \mathcal{A}$ and $t$ denotes time, and a reward function $\mathcal{R} : \mathcal{S} \times \mathcal{A} \rightarrow \mathbb{R}$.
It formulates a Markov Decision Process (MDP), represented as a tuple $(\mathcal{S}, \mathcal{A}, p, \mathcal{R})$.
A solution to an MDP is a policy $\pi$ which maps a state $s_t \in \mathcal{S}$ to a probability distribution over actions, i.e. $ a_t \sim \pi( \cdot | s_t )$.
The objective of Reinforcement Learning is to find an optimal policy $\pi^*$ that maximizes the expected discounted sum of rewards over a potentially infinite horizon, $\mathbb{E}\left[\sum _{t=0}^{\infty }{\gamma ^{t} \mathcal{R} (s_{t}, a_{t})}\right]$, where $\gamma$ is a discount factor.

When the task is parameterized by a goal, the goal is given as input to the policy.
Goal-conditioned policies, therefore, map from state $s_t \in \mathcal{S}$ and goal $g_t \in \mathcal{G}$, with the goal space $\mathcal{G}$, to a probability distribution over actions, i.e. $a_t \sim \pi( \cdot | s_t, g_t) $.
In a goal-conditioned formulation, the reward function is also goal-dependent, i.e.~$\mathcal{R}: \mathcal{S} \times \mathcal{A} \times \mathcal{G} \rightarrow \mathbb{R}$.

\subsection{Multiplicative Compositional Policy}
A Multiplicative Compositional Policy~\cite{Peng2019MCPLC} is a policy architecture that enables an agent to activate multiple primitives simultaneously, where each primitive specializes in a distinct behavior.
These behaviors can be composed to produce a continuous spectrum of skills. 
The policy enables that by treating $K$ primitives $\pi_{1}, ..., \pi_{K}$ as independent Gaussian distributions over actions.
The composite policy is obtained by a multiplicative composition of these distributions,
\begin{equation}
\label{eqn:multPrims}
\pi(a | s, g) = \frac{1}{Z(s, g)}\prod_{i = 1}^K \pi_i(a | s) ^{w_i(s, g)}, \quad w_i(s, g) \geq 0.
\end{equation}
A gating function $w$ specifies each weight $w_i(s, g) \in \mathbb{R}_{+}$, which determines the contribution of the $i$-th primitive on the composite action distribution, with a larger weight corresponding to a larger influence. 
$Z(s, g)$ is a normalizing term. 
The $i$-th primitive is modeled as a Gaussian distribution $\pi_i(a|s) = \mathcal{N}(\mu_i(s), \Sigma_i(s))$ with mean $\mu_i(s)$ and diagonal covariance matrix $\Sigma_i(s) = \mathrm{diag}\left(\sigma_i^1(s), \sigma_i^2(s), \ldots, \sigma_i^{|\mathcal{A}|}(s) \right)$, with the variance $\sigma_i^j(s)$ of the $j$-th action dimension from the $i$-th primitive, and the dimensionality of the action space $|\mathcal{A}|$. 
Each primitive $\pi_i(a|s)$ is independent of the goal $g$, which makes it reusable for any type of downstream task.
A multiplicative composition of Gaussian distributions yields yet another Gaussian distribution $\pi(a|s, g) = \mathcal{N}(\mu(s, g), \Sigma(s, g))$. 
Since the primitives model each dimension independently, the mean and variance of the composite policy $\pi$ also have independent dimensions with component-wise mean $\mu^j(s, g)$ and variance $\sigma^j(s, g)$ where
\begin{equation}
\mu^j(s, g) = \frac{1}{\sum_{l=1}^k \frac{w_l(s, g)}{\sigma_l^j(s)}} \sum_{i = 1}^k \frac{w_i(s, g)}{\sigma_i^j(s)} \mu_i^j(s),
\label{eqn:mcp_mu}
\end{equation}
and
\begin{equation}
\qquad \sigma^j(s, g) = \left(\sum_{i=1}^k \frac{w_i(s, g)}{\sigma_i^j(s)} \right)^{-1} .
\label{eqn:mcp_sigma}
\end{equation}
The gating function $w$ and the primitives $\pi_i$ mean and variance are parameterized by neural networks.
Hence the policy $\pi$ can be trained end-to-end using gradient descent.
\subsection{Asymmetric Self-Play}
Asymmetric Self-Play \cite{sukhbaatar2017intrinsic,sukhbaatar2018learning,OpenAI2021AsymmetricSF} is a curriculum learning approach that trains a goal \textit{generator}, Alice, along with a goal-conditioned \textit{solver}, Bob.
Alice starts from an arbitrary state $s_0$ in the environment and takes a sequence of actions that leads to some state $s_g$, which becomes the goal state of Bob.
The environment is then reset and Bob is tasked to reach the same state $s_g$ as Alice by starting from the same state $s_0$ that Alice started in.
Both agents are trained with RL.
On the one hand, Bob is rewarded for achieving the proposed goals.
On the other hand, to encourage the discovery of challenging and novel goals, Alice is rewarded for proposing goals that Bob is unable to achieve.
To prevent Alice from generating goals that are too complex for Bob to achieve, a penalty reward can also be given to Alice for proposing undesirable goals.
This penalty over goals can be hand-crafted to introduce a prior on the types of goals Alice must propose.
Moreover, Bob can always learn from Alice's trajectory to go from $s_0$ to $s_g$ and hence Bob always gets a positive learning signal even in cases of failure.
This adversarial reward structure yields a curriculum for Bob to learn from an increasingly difficult set of tasks.
We encourage the reader to refer to \cite{OpenAI2021AsymmetricSF} for a more detailed description of ASP.

\section{Method}
\label{sec:method}
We propose discovering diverse and composable primitives by pre-training an MCP to solve a large set of tasks generated through Asymmetry Self-Play.
This pre-training results in a collection of primitives that capture a range of useful behaviors explicitly learned to be composable in a manipulation setting.
The discovered behaviors can then be orchestrated to solve various (unseen) tasks in the environment using HRL.
We present an overview of our method in Figure~\ref{fig:method}.
\subsection{Discovering Behaviors through Tasks Diversity}
The core of our method is the pre-training part.
It requires a task \textit{generator} and a task \textit{solver}.
The role of the \textit{generator} is to propose tasks that the \textit{solver} must solve.
Our method relies on the hypothesis that if the set of tasks proposed by the \textit{generator} is large, diverse, and complex, then it induces the discovery of diverse and complex behaviors by the \textit{solver}.
While this idea is compatible with any automatic task generation framework that learns both a \textit{generator} and a \textit{solver}, we use Asymmetric Self-Play to discover a complex distribution of tasks.

We choose ASP for two main reasons.
Firstly, ASP is one of the few approaches that was shown to generate a diversity of complex tasks in a robotic manipulation setting~\cite{OpenAI2021AsymmetricSF}.
This consequence of the adversarial structure of ASP incites the \textit{generator}, Alice, to seek novel and complex goals.
Secondly, due to the structure of ASP, the tasks proposed by the \textit{generator} are guaranteed to be feasible by the \textit{solver}, Bob.
More importantly, each proposed task comes with a successful demonstration, which can be used to train the \textit{solver}, when it cannot succeed independently.
It is called Alice Behavioral Cloning (ABC).
It is of major importance in robotic manipulation to encourage proper interactions with objects.
Hence, we use it during pre-training.

Once the automatic task generation process terminates, i.e. when the \textit{generator} does not propose novel tasks anymore and the \textit{solver} can solve all proposed tasks, we keep the \textit{solver} for reuse with HRL on unseen downstream tasks.
We hypothesize that, as the \textit{solver} is exposed to a large variety of tasks, its policy embeds a diversity of behaviors necessary to solve all seen tasks.
We claim that these behaviors can be reused. 
\subsection{Embedding Behaviors in Composable Primitives}
We learn the \textit{solver} as a Multiplicative Compositional Policy to embed the whole range of discovered behaviors in primitives.
We parameterize the MCP with neural networks.
During pre-training, primitives learn to embed the discovered behaviors in their parameters while the orchestrator learns to compose the primitives to solve the proposed tasks.
Learning the solver's policy as a monolithic policy limits the range of options for downstream tasks.

Firstly, reusing a monolithic policy limits the range of downstream tasks to tasks with the same input space as in the pre-training phase.
For instance, if downstream tasks require input information that is not part of the input space of the \textit{solver} during pre-training, then the skills acquired during pre-training cannot be leveraged on downstream tasks.
MCPs address this problem by separately learning an orchestrator and a set of primitives.
Since the primitives only depend on the state $s$ and hence not on task-agnostic information, they can be transferred to any task, e.g., with different goal spaces.
The orchestrator can be learned while benefiting from the knowledge embedded in primitives.

Secondly, reusing a monolithic policy on downstream tasks can require fine-tuning.
It occurs when there is a significant shift between the pre-training and downstream task distributions or in the environment dynamics.
While fine-tuning can increase performance on some tasks, it can also induce catastrophic forgetting of the behaviors learned during pre-training.
A benefit of MCPs is that task-agnostic behaviors are embedded in individual primitives, which can be frozen after pre-training to prevent catastrophic forgetting.
The frozen primitives can then be composed to yield behaviors robust to changes in environment dynamics and task distribution shifts.
\subsection{Composing Primitives to Solve Downstream Tasks}
We hypothesize that if primitives can be composed to solve a sufficiently diverse set of tasks during pre-training, these primitives can then be repurposed in novel downstream tasks.
The primitives of the \textit{solver} are indeed pre-trained along with an orchestrator to solve the diverse tasks proposed by the \textit{generator}.
We hence have a guarantee that the primitives are effective and composable to solve all the tasks proposed by the \textit{generator}.
We hypothesize that since tasks proposed in the pre-training are diverse and complex, the primitives are general enough to serve a useful role in new tasks in the same environment.
We argue that a composition of diverse behaviors can adapt to various changes in the environment and facilitate learning new tasks.

This paper investigates the reusability of the learned primitives by an orchestrator trained from scratch.
For each downstream task, an orchestrator is trained from scratch to compose the pre-trained primitives.
To show how general and adaptable the learned primitives are, the very same set of pre-trained primitives is used on all downstream tasks, with frozen parameters.

\section{Experimental Setup}
\subsection{Tasks}
\begin{figure*}
    \centering
    \begin{subfigure}[b]{0.16\textwidth}
        \centering
        \includegraphics[width=0.99\textwidth]{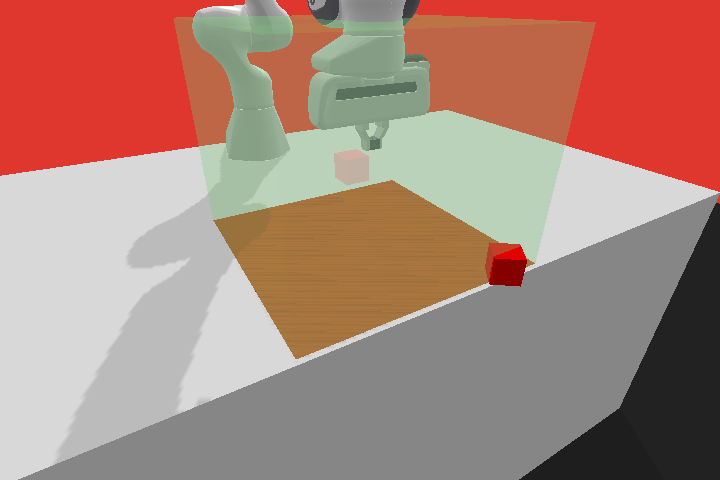}
        \caption{\textit{Larger}}
    \end{subfigure}
    \begin{subfigure}[b]{0.16\textwidth}
        \centering
        \includegraphics[width=0.99\textwidth]{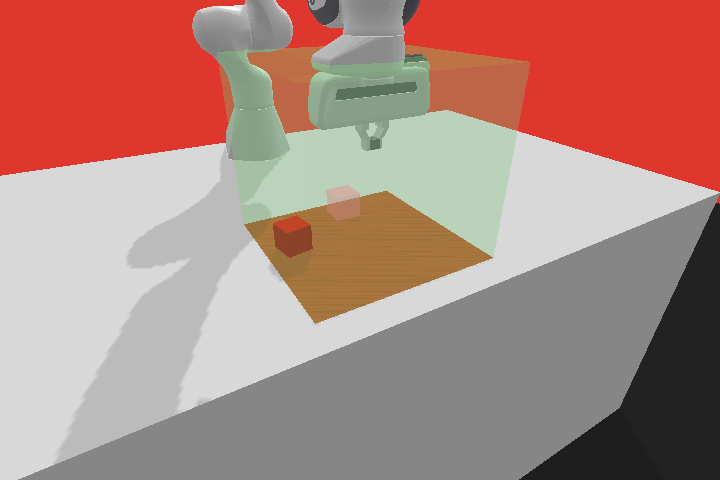}
        \caption{\textit{In Air}}
    \end{subfigure}
    \begin{subfigure}[b]{0.16\textwidth}
        \centering
        \includegraphics[width=0.99\textwidth]{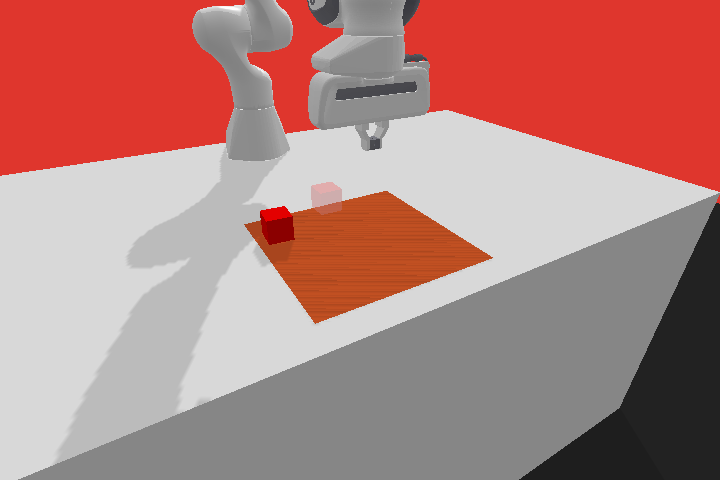}
        \caption{\textit{Push}}
    \end{subfigure}
    \begin{subfigure}[b]{0.16\textwidth}
        \centering
        \includegraphics[width=0.99\textwidth]{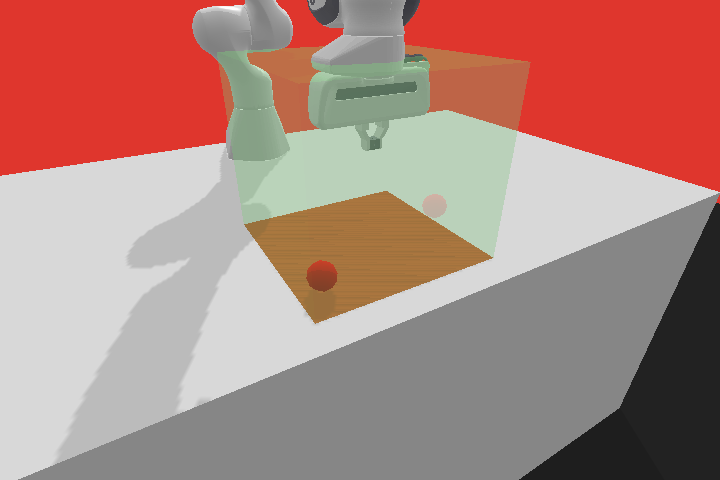}
        \caption{\textit{Sphere}}
    \end{subfigure}
    \begin{subfigure}[b]{0.16\textwidth}
        \centering
        \includegraphics[width=0.99\textwidth]{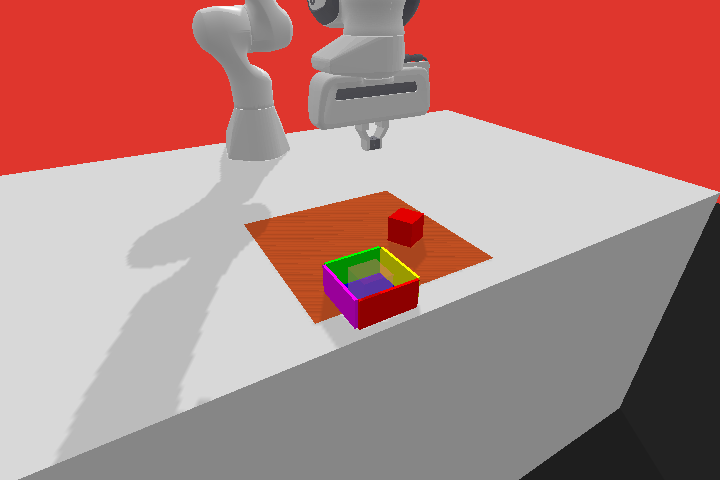}
        \caption{\textit{Box}}
    \end{subfigure}
    \begin{subfigure}[b]{0.16\textwidth}
        \centering
        \includegraphics[width=0.99\textwidth]{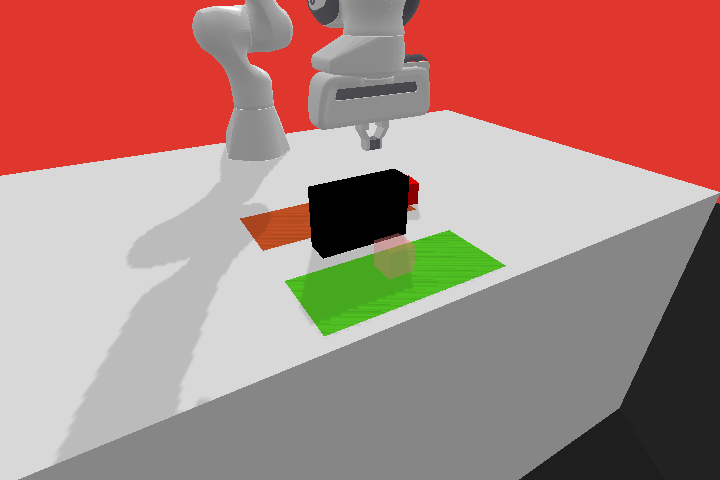}
        \caption{\textit{Wall}}
    \end{subfigure}
    \caption{\textbf{Downstream Tasks}:
    The robot must move the solid red cube to a target position represented as a transparent red cube.
    The orange zone represents where the initial position of the cube can be sampled, while the green zone represents where the target position of the cube can be sampled.
    }
    \label{fig:downstream_tasks}
\end{figure*}
\label{sec:setup}
We apply our method in a customized version of \texttt{panda-gym}~\cite{gallouedec2021pandagym} for pre-training and downstream task training.
The environment consists of a robotic arm placed in a table-top setting where an object can be manipulated.
Similar environments are commonly used to evaluate Skill Learning methods for Robotic Manipulation~\cite{park2021lipschitz,pmlr-v202-park23h}.
We use a Franka Emika Panda with a parallel gripper in all experiments.
We focus on diverse manipulation tasks involving a single rigid object.
Pre-training and downstream training are done in simulation only but we evaluate the transferability of our primitives to a real robotic platform on two downstream tasks.
The action space corresponds to the displacement of the gripper and its finger.
The state space is composed of the state of the robot and of the object.
The reward function is a sparse success-based reward.
The tasks consist of moving an object to a goal position.
The task is considered solved when the object has achieved its target position.
We consider an object has achieved its goal position if it is within a distance $d_\text{threshold} = 5$~cm from it.
While \cite{OpenAI2021AsymmetricSF} showed that ASP could be extended to tasks involving both the position and orientation of objects, we limit our study to position only as it is enough to show the benefits of our method.

During pre-training, Alice and Bob manipulate a $5$~cm cube.
Alice proposes goals in the workspace and is penalized otherwise.
We define the workspace as a volume of dimensions $35 \times 35 \times 35$~cm, on the table.
The initial object position is uniformly sampled in the workspace on the table.

Downstream tasks are variations of pick-and-place tasks.
We use the same workspace as in pre-training and sample initial object position similarly.
Goal positions are uniformly sampled anywhere in the workspace with a probability $p$ and otherwise on the table.
The manipulated object is a cube of size $5$~cm unless specified otherwise.
Each downstream task perturbs the dynamics of the environment or the task distribution w.r.t. to pre-training to evaluate how a composition of pre-trained primitives can adapt.
In \textit{Larger}, the area where initial and target object positions are sampled is of size $45 \times 45 \times 40$~cm and $p=0.7$.
In \textit{In Air}, target positions are always sampled in the air, i.e. $p=1.0$,
For \textit{Push}, the gripper is blocked in a closed position, and targets are always on the table, i.e., $p=0.0$,
In Sphere \textit{Sphere}, the object is a sphere of diameter $D=5$~cm and $p=0.7$,
For \textit{Wall}, the initial and target object positions are sampled from disjoint areas of size $15 \times 35 \times 0$~cm separated by a fixed wall, and targets are always on the table, i.e. $p=0.7$,
In \textit{Box}, a rigid open box of size $10 \times 10 \times 5$~cm is welded at a random position to the table, and the goal is inside the box ($p=0.0$).
\subsection{Baselines}
We compare our approach to the following baselines:
(a) LSD~\cite{park2021lipschitz} with object position prior: we train the state representation function to transform the position of the object such that its transition is aligned with a latent descriptor $z$;
(b) DADS~\cite{Sharma2020DynamicsAwareUS} with object position prior: we train the dynamics predictor to predict the next position of the object given the current position and a continuous skill latent descriptor $z$;
(c) DIAYN~\cite{eysenbach2019_diayn} with object position prior: we train the skill predictor to predict the continuous skill latent descriptor $z$ given the position of the object;
(d) ASP~\cite{OpenAI2021AsymmetricSF} only: we use Bob's policy from ASP pre-training only, i.e., without any training on downstream tasks;
(e) Scratch MCP with Hindsight Experience Replay (HER)~\cite{andrychowicz2017hindsight}: we train an MCP from scratch with HER;
(f) Scratch MCP without HER: we train an MCP from scratch without HER;
(g) Scratch Monolithic with HER: we train a monolithic policy from scratch with HER;
(h) Scratch Monolithic without HER: we train a monolithic policy from scratch without HER.
On downstream tasks, skills pre-trained with LSD, DADS, DIAYN, and our approach are reused with HRL.
Scratch baselines are trained end-to-end and do not benefit from pre-training.

\section{Evaluation}
\label{sec:results}
We want to answer the following questions:
(1)~How do our primitives interact with objects?
(2)~Can our primitives be efficiently repurposed for new tasks?
(3)~What are the benefits of our approach compared to ASP and MCP used in isolation?
(4)~Can our primitives be reused in a real-world setting?

\begin{figure*}
    \centering
    \begin{subfigure}[b]{0.49\textwidth}
        \centering
        \includegraphics[width=0.325\columnwidth]{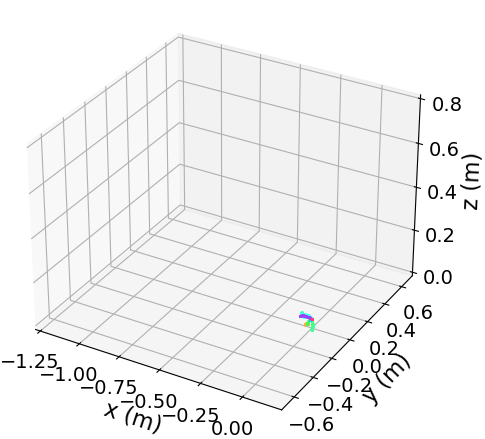}
        \includegraphics[width=0.325\columnwidth]{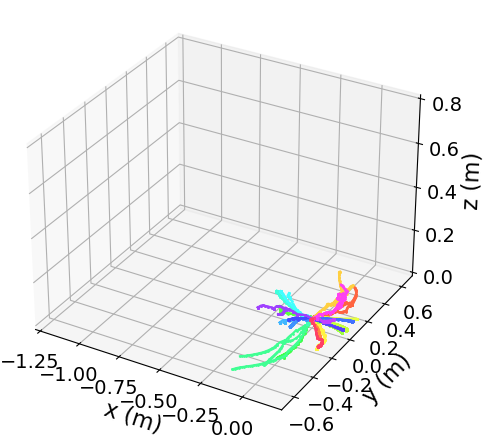}
        \includegraphics[width=0.325\columnwidth]{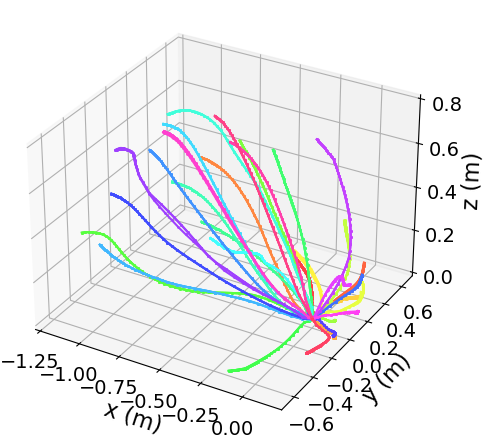}
        \caption{Ours}
        \label{fig:coverage_ours}
    \end{subfigure}
    \begin{subfigure}[b]{0.16\textwidth}
        \centering
        \includegraphics[width=\columnwidth]{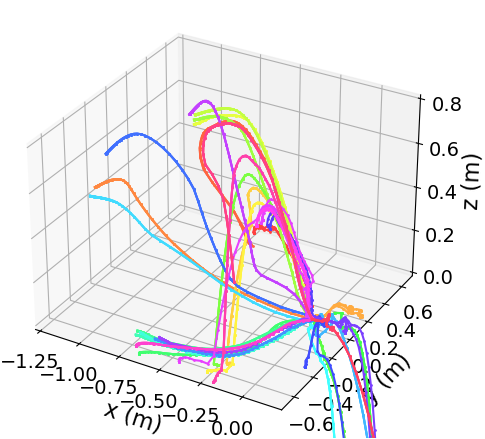}
        \caption{LSD}
        \label{fig:coverage_lsd}
    \end{subfigure}
    \begin{subfigure}[b]{0.16\textwidth}
        \centering
        \includegraphics[width=\columnwidth]{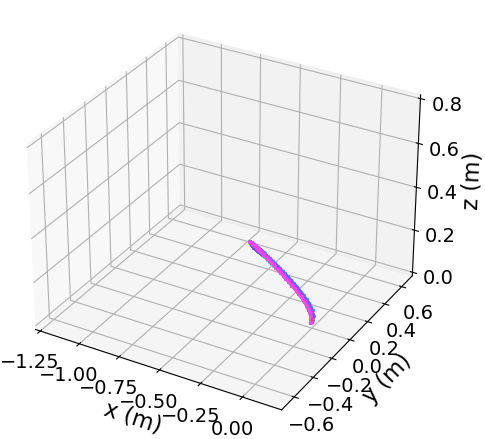}
        \caption{DADS}
        \label{fig:coverage_dads}
    \end{subfigure}
    \begin{subfigure}[b]{0.16\textwidth}
        \centering
        \includegraphics[width=\columnwidth]{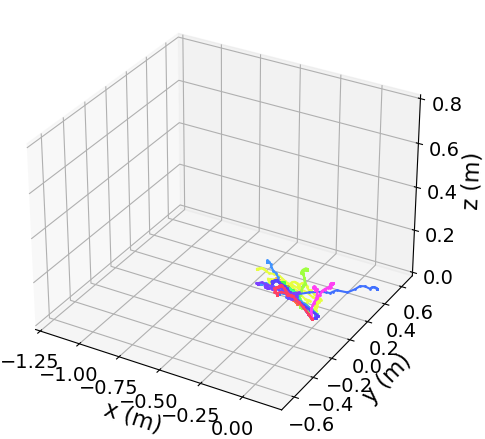}
        \caption{DIAYN}
        \label{fig:coverage_diayn}
    \end{subfigure}
    \caption{\textbf{Per-Skill Trajectories}: 
    Trajectories of object positions for $N_\text{skills}=\text{32}$ random compositions of skills.
    Each color corresponds to a random composition.
    Our primitives in \ref{fig:coverage_ours} increasingly learn to cover reachable positions of the object.
    The baselines partially cover the workspace (\ref{fig:coverage_diayn},\ref{fig:coverage_lsd}), push and throw the object to unreachable positions (\ref{fig:coverage_lsd}), or very rarely interact with the object (\ref{fig:coverage_dads}).}
    \label{fig:coverage}
\end{figure*}
\subsection{Primitives Progression through Pre-Training}
\label{subsec:primitives_progression_coverage}
We examine the progression of our learned primitives throughout pre-training and evaluate their coverage in terms of object positions.
We believe that analyzing compositions of primitives provides more insights into their behaviors than analyzing individual primitives alone. 
During pre-training, we qualitatively assess the evolution of our primitives by keeping checkpoints at three stages.
We sample $N_{\text{skill}}$ random values for the gate $w \in \mathbb{R}^K$, which determines a fixed composition of primitives. %

As shown in Figure~\ref{fig:coverage}, the primitives do not exhibit meaningful behaviors in early pre-training, resulting in minimal object movement.
However, as pre-training progresses and proposed tasks require moving the object on the table, the primitives learn to interact with the object.
Subsequently, as tasks requiring object lifting are introduced, the primitives learn to exhibit lifting behaviors.
Towards the end of pre-training, after the robot has been exposed to all tasks, the learned primitives demonstrate the ability to move the object to arbitrary positions in the workspace.
This observation suggests that task diversity plays a crucial role in discovering new behaviors, and these behaviors are effectively embedded in the primitives.

To evaluate the coverage of our primitives, we compare them to skills learned with LSD, DADS, and DIAYN.
Following a similar protocol, we save the skillset only at the end of pre-training.
For baselines, we sample $N_{\text{skill}}$ fixed random values of the latent skill descriptor $z$, which are then given as fixed inputs to the skill encoder.
The results of our evaluation, depicted in Figure~\ref{fig:coverage}, reveal that our method achieves full coverage of reachable object positions.
Compositions of our primitives indeed span all reachable object positions, ensuring the ability to move the object anywhere within the workspace.
Meanwhile, LSD learns skills that partially cover the space of object positions but also push and throw the object in unreachable areas, such as off the table.
We believe this is a direct consequence of the intrinsic objective of LSD, which mainly encourages the discovery of dynamic behaviors.
This lack of coverage and tendency to move objects off the table can be detrimental to solving downstream tasks in a table-top setting.
Our method learns behaviors that are useful for tasks in a table-top setting, which encourages coverage while discouraging moving the object to irrecoverable areas.
The skills learned with DADS and DIAYN exhibit few interactions with the object.
In particular, DIAYN learns skills that achieve a low coverage of object positions while DADS skillset collapses to either ignoring the object or moving it along a fixed path.
These limitations are likely due to the sparsity and noise of their intrinsic rewards in robotic manipulation.
Attempts to address this by adding bonuses to the intrinsic rewards did not result in the learning of interactive skills.
The limited coverage of object positions achieved by DADS and DIAYN restricts the reusability of these skills in downstream tasks.
In contrast, our method overcomes this limitation thanks to the curriculum induced by ASP and ABC, and the explicit design to promote interactions with objects.
\subsection{Primitives Orchestration on Downstream Tasks}
\begin{figure*}
    \centering
    \includegraphics[width=\textwidth]{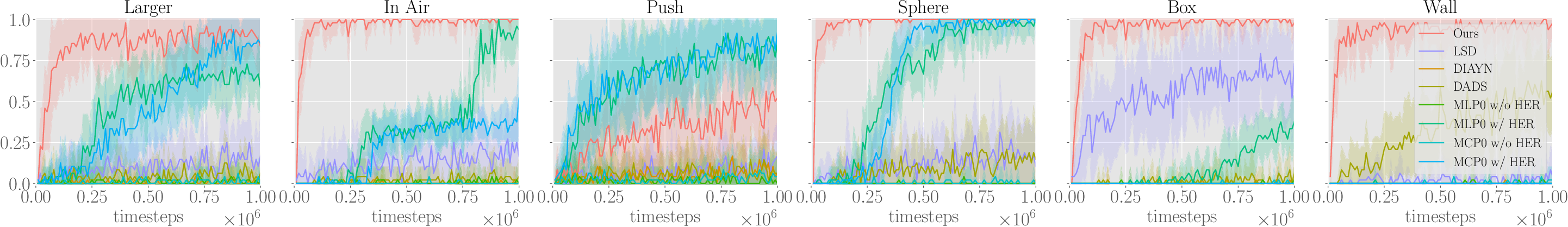}
    \caption{\textbf{Downstream Learning Curves}:
    Average success rate throughout training on a downstream task.
    The solid area represents the standard deviation.
    "Scratch" baselines are denoted with a "0" for compactness.
    Our method is overall the most competitive both in terms of sample efficiency and final performance.
    }
    \label{fig:tasks}
\end{figure*}
\begin{table}
\setlength{\tabcolsep}{2.8pt}.
    \centering
    \begin{tabular}{lcccccc}
        \toprule
         & \textit{Larger} & \textit{In Air} & \textit{Push} & \textit{Sphere} & \textit{Wall} & \textit{Box} \\
        \midrule
        Ours (simulation)   & 0.83  & \textbf{1.00}  & 0.52  & \textbf{1.00}  & \textbf{0.98}  & \textbf{0.98} \\
        ASP only            & \textbf{0.94}  & 0.94  & 0.02  & 0.85  & 0.00  & 0.27 \\
        LSD          & 0.17  & 0.17  & 0.17     & 0.10  & 0.04  & 0.73 \\
        DADS                & 0.06  & 0.04  & 0.06     & 0.15  & 0.56  & 0.06 \\
        DIAYN                & 0.00  & 0.00  & 0.06    & 0.02  & 0.00  & 0.02 \\
        Scratch Monolithic w/ HER  & 0.58  & 0.94  & 0.85  & \textbf{1.00}  & 0.00  & 0.36 \\
        Scratch MCP w/ HER  & 0.85  & 0.52  & \textbf{0.92}  & 0.98  & 0.00  & 0.00 \\
        Scratch Monolithic w/o HER & 0.02  & 0.00  & 0.00  & 0.00  & 0.00  & 0.00 \\
        Scratch MCP w/o HER & 0.00  & 0.00  & 0.04  & 0.00  & 0.02  & 0.00 \\
        \midrule
        Ours (real)         & 0.83  &   -   &   -   &   -   & 0.59  &  -   \\
        \bottomrule
    \end{tabular}
    \caption{
    \textbf{Downstream Success Rates}: Each success rate is computed after training on each downstream task and averaged over three random seeds.
    }
    \label{tab:results}
\end{table}

We train orchestrators from scratch in simulation and evaluate how they learn to compose our pre-trained primitives to solve unseen tasks.
For each task, we train a different orchestrator while using the same set of pre-trained primitives with frozen parameters.
In Figure~\ref{fig:tasks}, we show the evolution of the success rate on these tasks for our method and baselines averaged over three random seeds.
In our accompanying video, we additionally provide visualizations of each Skill Learning agent on downstream tasks.
Table~\ref{tab:results} reports the final achieved success rate.
Most Skill Learning baselines and agents trained from scratch without HER fail on all tasks.
On \textit{Larger}, \textit{In Air}, and \textit{Sphere}, our orchestrators learn faster than all baselines, including agents benefiting from HER, and reach the highest success rates.
This suggests that our primitives can be reused on tasks they were not explicitly pre-trained to solve and that they can achieve high performance with high sample efficiency.
We find that LSD achieves limited performance on these three tasks, and we qualitatively observe that agents reusing skills learned with LSD can only reach goals in some areas of the workspace and do not stabilize to the target position but rather oscillate around it.
In \textit{Sphere}, LSD skills often fail to grasp the object stably. %
We suggest that this is a consequence of LSD encouraging the discovery of dynamic skills, which makes it difficult to maintain the object at a static position.
DADS and DIAYN also achieve low success rates on these tasks.
Despite the changes in dynamics induced by a new object in the environment, our approach quickly achieves high success rates on \textit{Wall} and \textit{Box}.
Surprisingly, LSD skills achieve reasonable performance on \textit{Box} despite their limited results on other tasks.
We find that the oscillations of the end-effector around the target position that we observed in other tasks are alleviated by the box, which constrains the motion of the end-effector to remain inside the box.
Hence, in this very specific scenario, part of the issues induced by dynamic skills are addressed by the structure of the environment.
On \textit{Wall}, LSD skills fail to solve the tasks, and we believe that this is due to the lack of coverage of the skillset.
On \textit{Box}, both DIAYN and DADS fail.
However, on \textit{Wall}, DADS obtains a surprisingly high score.
We qualitatively observe that DADS always solves the task in a suboptimal manner by using the unique interactive behavior it discovered, by following a long trajectory to avoid the obstacle.
Meanwhile, all other approaches get low success rates, if any.
This shows that our primitives can be composed and used in settings with different dynamics they were pre-trained on.
\textit{Push} is where our approach struggles the most, while HER performs well.
As the pre-trained primitives favor grasping, a blocked gripper makes them significantly less efficient in manipulating the object.
LSD skills seem to suffer even more from this issue as its success rate on \textit{Push} is significantly lower.
This illustrates a change in dynamics where no Skill Learning method can adapt.
From Table~\ref{tab:results}, we observe that ASP in isolation does not achieve as good results on most tasks, especially with perturbations to the dynamics.
We also observe that MCP trained from scratch in isolation performs similarly to its monolithic counterpart, if not worse.
We think this is due to the more unstable nature of MCP.
Overall, these results suggest that the benefits of our method do not come from ASP or MCP in isolation but from their combination, yielding the discovery of adaptable behaviors.

We evaluate the primitives in the real world and show the real robot using our primitives in our accompanying video.
We track the position of the object with a motion capture system.
We test the models trained in simulation to solve \textit{Large Area} and \textit{Wall}, without additional training.
As we found that the learned policies tend to push the gripper against the table, we clip the action dimension of the gripper such that its position always remains slightly above the level of the table, hence avoiding collisions.
We found this to be strictly necessary to run our trained policies on the robot safely.
We diversify the initial and target positions such that the object must move in all regions of the workspace.
We run 103 trials and 36 for \textit{Larger} and \textit{Wall}, respectively.
While the wall is solid and fixed in simulation, it is made of cardboard in our real-world setup to avoid damaging the robot in case of collisions.
We report the average success rate of both tasks in Table~\ref{tab:results}.
On the one hand, the model trained to solve \textit{Larger} performs as well as in simulation.
Most failures come from target positions that are at the edge of the reachable area of the end-effector, which triggers failures of the controller.
This suggests that the primitives learned robust behaviors that can be transferred to real-world settings.
On the other hand, the model trained to solve \textit{Wall} does not perform as well.
Most failures occur when the target is at the edge of the goal area or when the end-effector collides with the wall.
This is a consequence of the sparse reward, which only encourages success without safety considerations.
Overall, these results are encouraging as they show our primitives can be used on a real robot to solve manipulation tasks.

\section{Conclusion}

This paper introduces a novel approach to learning composable and reusable skills in robotic manipulation. 
We propose to leverage Asymmetric Self-Play to generate diverse and complex pre-training tasks and Multiplicative Compositional Policies to obtain a skill repertoire for diverse object manipulation tasks.
We show the effectiveness of our method in simulated and real-world robotic manipulation scenarios on a set of downstream tasks with Hierarchical Reinforcement Learning.
While showing promising results, some limitations still need to be addressed.
A limitation of our approach is the object-centric observations and we would like to tackle Skill Learning from visual inputs in future work.
Other future directions lie in finding better model architectures to embed and reuse the discovered behaviors.

\section{Appendix}

\subsection{Environment}
\label{apx:environment}

\subsubsection{State}
The state of the environment consists of the end-effector position and linear velocity, the finger positions and linear velocities, the object absolute positions, the object position relative to the end-effector, the object linear velocity, binary contact information between fingers and the object, and the number of steps taken since the beginning of the episode.
Alice's policy receives the full state as input.
We find that including contact information and time in Alice's inputs facilitates the discovery of novel tasks and behaviors, such as grasping objects.
Since Alice is only a proxy to train Bob and is never deployed on the real robot, she can benefit from privileged information as input.
As contact and velocity information would be too different between simulation and reality, we do not provide these to Bob.
Bob's primitives and its orchestrator only receive positions as inputs.
In addition, as in Eq.~\ref{eqn:multPrims}, the orchestrator also takes task-specific information as input, which consists of the target object position in our experiments.
It ensures that primitives only depend on reliable information in the real robotic platform, making the primitives more transferable to the real world.
During downstream training, all agents receive only positions and goals as input.
\subsubsection{Action}
The action is four-dimensional, comprising both the desired displacement of the end-effector and the change in width between the fingers.
Actions are scaled from $[-1, +1]$ to $[-5, +5]$ cm for the end-effector and to $[-10, +10]$ cm for finger displacements.
Reference end-effector positions are obtained by applying the displacements to the current positions.
During downstream training, the size of the action space of the orchestrator is the number of primitives or the size of the latent space for Skill Learning baselines. %
\subsubsection{Reward}
We use sparse rewards.
During pre-training, we use a comparable adversarial reward structure as used in~\cite{OpenAI2021AsymmetricSF}.
On the one hand, Bob is rewarded for solving the task at the moment the task is solved.
On the other hand, Alice is rewarded at the end of each episode and gets a reward $r_{\text{valid}} = +1$ if it proposes a valid goal.
A goal position is valid if it is defined in the workspace, and if such position is different from the initial position of the object.
If the goal is valid, Alice gets an additional reward $r_{\text{difficult}} = 5$ if Bob fails to solve the task or $r_{\text{difficult}} = 0$ if Bob succeeds.
During downstream training, the agent is rewarded at every step where an object is on its target.

\subsection{Implementation Details}

For pre-training, we use Proximal Policy Optimization (PPO)~\cite{schulman2017ppo} to update Alice's and Bob's policies as in~\cite{OpenAI2021AsymmetricSF}.
We re-implemented PPO as we needed to articulate it with ABC for Bob.
For downstream training, we use the implementation of Soft Actor-Critic~\cite{haarnoja2018soft} from Stable Baselines3~\cite{stable-baselines3}, for a higher sample efficiency, and compatibility with HER.

To ensure a fair comparison to Skill Learning methods, we set the dimensionality of their skill latent space equal to the number of primitives $K$.
We use $K=4$ for all experiments as we did not find that a higher number helped significantly on the evaluated tasks.

\section*{ACKNOWLEDGMENT}
We thank Jean-Michel Renders, David Emukpere, and Seungsu Kim for their thoughtful feedback. 
\bibliographystyle{IEEEtran}
\bibliography{bibliography}  %

\begin{thebibliography}{10}
\providecommand{\url}[1]{#1}
\csname url@rmstyle\endcsname
\providecommand{\newblock}{\relax}
\providecommand{\bibinfo}[2]{#2}
\providecommand\BIBentrySTDinterwordspacing{\spaceskip=0pt\relax}
\providecommand\BIBentryALTinterwordstretchfactor{4}
\providecommand\BIBentryALTinterwordspacing{\spaceskip=\fontdimen2\font plus
\BIBentryALTinterwordstretchfactor\fontdimen3\font minus
  \fontdimen4\font\relax}
\providecommand\BIBforeignlanguage[2]{{%
\expandafter\ifx\csname l@#1\endcsname\relax
\typeout{** WARNING: IEEEtran.bst: No hyphenation pattern has been}%
\typeout{** loaded for the language `#1'. Using the pattern for}%
\typeout{** the default language instead.}%
\else
\language=\csname l@#1\endcsname
\fi
#2}}

\bibitem{riedmiller2018sacX}
M.~Riedmiller, R.~Hafner, T.~Lampe, M.~Neunert, J.~Degrave, T.~V. de~Wiele,
  V.~Mnih, N.~Heess, and J.~T. Springenberg, ``Learning by playing - solving
  sparse reward tasks from scratch,'' 2018.

\bibitem{vecerik2018robotics}
M.~Vecerik, T.~Hester, J.~Scholz, F.~Wang, O.~Pietquin, B.~Piot, N.~Heess,
  T.~Rothörl, T.~Lampe, and M.~Riedmiller, ``Leveraging demonstrations for
  deep reinforcement learning on robotics problems with sparse rewards,'' 2018.

\bibitem{popov2017lego}
I.~Popov, N.~Heess, T.~Lillicrap, R.~Hafner, G.~Barth-Maron, M.~Vecerik,
  T.~Lampe, Y.~Tassa, T.~Erez, and M.~Riedmiller, ``Data-efficient deep
  reinforcement learning for dexterous manipulation,'' \emph{arXiv preprint
  arXiv:1704.03073}, 2017.

\bibitem{yu2020meta}
T.~Yu, D.~Quillen, Z.~He, R.~Julian, K.~Hausman, C.~Finn, and S.~Levine,
  ``Meta-world: A benchmark and evaluation for multi-task and meta
  reinforcement learning,'' in \emph{Conference on Robot Learning}, 2020.

\bibitem{andrychowicz2017hindsight}
M.~Andrychowicz, F.~Wolski, A.~Ray, J.~Schneider, R.~Fong, P.~Welinder,
  B.~McGrew, J.~Tobin, P.~Abbeel, and W.~Zaremba, ``Hindsight experience
  replay,'' in \emph{Advances in neural information processing systems}, 2017.

\bibitem{nair2018demo}
A.~Nair, B.~McGrew, M.~Andrychowicz, W.~Zaremba, and P.~Abbeel, ``Overcoming
  exploration in reinforcement learning with demonstrations,'' in \emph{IEEE
  International Conference on Robotics and Automation (ICRA)}, 2018.

\bibitem{eysenbach2019_diayn}
B.~Eysenbach, A.~Gupta, J.~Ibarz, and S.~Levine, ``Diversity is all you need:
  Learning skills without a reward function,'' in \emph{ICLR}, 2019.

\bibitem{Sharma2020DynamicsAwareUS}
A.~Sharma, S.~S. Gu, S.~Levine, V.~Kumar, and K.~Hausman, ``Dynamics-aware
  unsupervised skill discovery,'' in \emph{ICLR 2020}, 2020.

\bibitem{achiam2018_valor}
J.~Achiam, H.~Edwards, D.~Amodei, and P.~Abbeel, ``Variational option discovery
  algorithms,'' \emph{arXiv preprint arXiv:1807.10299}, 2018.

\bibitem{gregor2016_vic}
K.~Gregor, D.~J. Rezende, and D.~Wierstra, ``Variational intrinsic control,''
  \emph{arXiv preprint arXiv:1611.07507}, 2016.

\bibitem{warde2019_discern}
D.~Warde-Farley, T.~Van~de Wiele, T.~Kulkarni, C.~Ionescu, S.~Hansen, and
  V.~Mnih, ``Unsupervised control through non-parametric discriminative
  rewards,'' in \emph{ICLR}, 2019.

\bibitem{pong2019_skewfit}
V.~H. Pong, M.~Dalal, S.~Lin, A.~Nair, S.~Bahl, and S.~Levine, ``Skew-fit:
  State-covering self-supervised reinforcement learning,'' \emph{arXiv preprint
  arXiv:1903.03698}, 2019.

\bibitem{park2021lipschitz}
S.~Park, J.~Choi, J.~Kim, H.~Lee, and G.~Kim, ``Lipschitz-constrained
  unsupervised skill discovery,'' in \emph{International Conference on Learning
  Representations}, 2021.

\bibitem{pmlr-v202-park23h}
\BIBentryALTinterwordspacing
S.~Park, K.~Lee, Y.~Lee, and P.~Abbeel, ``Controllability-aware unsupervised
  skill discovery,'' in \emph{Proceedings of the 40th International Conference
  on Machine Learning}, ser. Proceedings of Machine Learning Research,
  A.~Krause, E.~Brunskill, K.~Cho, B.~Engelhardt, S.~Sabato, and J.~Scarlett,
  Eds., vol. 202.\hskip 1em plus 0.5em minus 0.4em\relax PMLR, 23--29 Jul 2023,
  pp. 27\,225--27\,245. [Online]. Available:
  \url{https://proceedings.mlr.press/v202/park23h.html}
\BIBentrySTDinterwordspacing

\bibitem{laskin2022unsupervised}
M.~Laskin, H.~Liu, X.~B. Peng, D.~Yarats, A.~Rajeswaran, and P.~Abbeel,
  ``Unsupervised reinforcement learning with contrastive intrinsic control,''
  \emph{Advances in Neural Information Processing Systems}, vol.~35, pp.
  34\,478--34\,491, 2022.

\bibitem{campos2020explore}
V.~Campos, A.~Trott, C.~Xiong, R.~Socher, X.~Gir{\'o}-i Nieto, and J.~Torres,
  ``Explore, discover and learn: Unsupervised discovery of state-covering
  skills,'' in \emph{International Conference on Machine Learning}.\hskip 1em
  plus 0.5em minus 0.4em\relax PMLR, 2020, pp. 1317--1327.

\bibitem{sukhbaatar2017intrinsic}
S.~Sukhbaatar, Z.~Lin, I.~Kostrikov, G.~Synnaeve, A.~Szlam, and R.~Fergus,
  ``Intrinsic motivation and automatic curricula via asymmetric self-play,'' in
  \emph{International Conference on Learning Representations}, 2018.

\bibitem{sukhbaatar2018learning}
S.~Sukhbaatar, E.~Denton, A.~Szlam, and R.~Fergus, ``Learning goal embeddings
  via self-play for hierarchical reinforcement learning,'' \emph{arXiv preprint
  arXiv:1811.09083}, 2018.

\bibitem{OpenAI2021AsymmetricSF}
O.~OpenAI, M.~Plappert, R.~Sampedro, T.~Xu, I.~Akkaya, V.~Kosaraju,
  P.~Welinder, R.~D'Sa, A.~Petron, H.~P. de~Oliveira~Pinto, A.~Paino, H.~Noh,
  L.~Weng, Q.~Yuan, C.~Chu, and W.~Zaremba, ``Asymmetric self-play for
  automatic goal discovery in robotic manipulation,'' \emph{ArXiv}, vol.
  abs/2101.04882, 2021.

\bibitem{Peng2019MCPLC}
X.~B. Peng, M.~Chang, G.~H. Zhang, P.~Abbeel, and S.~Levine, ``Mcp: Learning
  composable hierarchical control with multiplicative compositional policies,''
  in \emph{NeurIPS}, 2019.

\bibitem{stolle2002learning}
M.~Stolle and D.~Precup, ``Learning options in reinforcement learning,'' in
  \emph{Abstraction, Reformulation, and Approximation: 5th International
  Symposium, SARA 2002 Kananaskis, Alberta, Canada August 2--4, 2002
  Proceedings 5}.\hskip 1em plus 0.5em minus 0.4em\relax Springer, 2002, pp.
  212--223.

\bibitem{bacon2017option}
P.-L. Bacon, J.~Harb, and D.~Precup, ``The option-critic architecture,'' in
  \emph{Proceedings of the AAAI conference on artificial intelligence},
  vol.~31, no.~1, 2017.

\bibitem{DBLP:journals/corr/abs-2106-13105}
\BIBentryALTinterwordspacing
A.~Barreto, D.~Borsa, S.~Hou, G.~Comanici, E.~Ayg{\"{u}}n, P.~Hamel, D.~Toyama,
  J.~J. Hunt, S.~Mourad, D.~Silver, and D.~Precup, ``The option keyboard:
  Combining skills in reinforcement learning,'' \emph{CoRR}, vol.
  abs/2106.13105, 2021. [Online]. Available:
  \url{https://arxiv.org/abs/2106.13105}
\BIBentrySTDinterwordspacing

\bibitem{cho2022unsupervised}
D.~Cho, J.~Kim, and H.~J. Kim, ``Unsupervised reinforcement learning for
  transferable manipulation skill discovery,'' \emph{IEEE Robotics and
  Automation Letters}, vol.~7, no.~3, pp. 7455--7462, 2022.

\bibitem{florensa2017reverse}
C.~Florensa, D.~Held, M.~Wulfmeier, M.~Zhang, and P.~Abbeel, ``Reverse
  curriculum generation for reinforcement learning,'' in \emph{Conference on
  Robot Learning}, 2017.

\bibitem{salimans2018backward}
T.~Salimans and R.~Chen, ``Learning montezuma's revenge from a single
  demonstration,'' \emph{arXiv preprint arXiv:1812.03381}, 2018.

\bibitem{matiisen2019TSCL}
T.~Matiisen, A.~Oliver, T.~Cohen, and J.~Schulman, ``Teacher-student curriculum
  learning,'' \emph{IEEE transactions on neural networks and learning systems},
  2019.

\bibitem{zhang2020automatic}
Y.~Zhang, P.~Abbeel, and L.~Pinto, ``Automatic curriculum learning through
  value disagreement,'' \emph{arXiv preprint arXiv:2006.09641}, 2020.

\bibitem{portelas2020teacher}
R.~Portelas, C.~Colas, K.~Hofmann, and P.-Y. Oudeyer, ``Teacher algorithms for
  curriculum learning of deep rl in continuously parameterized environments,''
  in \emph{Conference on Robot Learning}.\hskip 1em plus 0.5em minus
  0.4em\relax PMLR, 2020, pp. 835--853.

\bibitem{oudeyer2007lp}
P.-Y. Oudeyer, F.~Kaplan, and V.~V. Hafner, ``Intrinsic motivation systems for
  autonomous mental development,'' \emph{IEEE transactions on evolutionary
  computation}, vol.~11, no.~2, pp. 265--286, 2007.

\bibitem{baranes2013motivation}
A.~Baranes and P.-Y. Oudeyer, ``Active learning of inverse models with
  intrinsically motivated goal exploration in robots,'' \emph{Robotics and
  Autonomous Systems}, vol.~61, no.~1, pp. 49--73, 2013.

\bibitem{pathak2017curiosity}
D.~Pathak, P.~Agrawal, A.~A. Efros, and T.~Darrell, ``Curiosity-driven
  exploration by self-supervised prediction,'' in \emph{Proceedings of the 34th
  International Conference on Machine Learning-Volume 70}, 2017.

\bibitem{burda2018exploration}
Y.~Burda, H.~Edwards, A.~Storkey, and O.~Klimov, ``Exploration by random
  network distillation,'' in \emph{International Conference on Learning
  Representations}, 2019.

\bibitem{ecoffet2019GoExplore}
A.~Ecoffet, J.~Huizinga, J.~Lehman, K.~O. Stanley, and J.~Clune, ``Go-explore:
  a new approach for hard-exploration problems,'' \emph{arXiv preprint
  arXiv:1901.10995}, 2019.

\bibitem{Ecoffet2020FirstRT}
\BIBentryALTinterwordspacing
A.~Ecoffet, J.~Huizinga, J.~Lehman, K.~O. Stanley, and J.~\vspace{0mm}Clune,
  ``First return, then explore,'' \emph{Nature}, vol. 590, pp. 580 -- 586,
  2020. [Online]. Available:
  \url{https://api.semanticscholar.org/CorpusID:216552951}
\BIBentrySTDinterwordspacing

\bibitem{oh2018SIL}
J.~Oh, Y.~Guo, S.~Singh, and H.~Lee, ``Self-imitation learning,'' \emph{arXiv
  preprint arXiv:1806.05635}, 2018.

\bibitem{wang2019poet}
R.~Wang, J.~Lehman, J.~Clune, and K.~O. Stanley, ``Paired open-ended
  trailblazer (poet): Endlessly generating increasingly complex and diverse
  learning environments and their solutions,'' \emph{arXiv preprint
  arXiv:1901.01753}, 2019.

\bibitem{wang2020enhanced}
R.~Wang, J.~Lehman, A.~Rawal, J.~Zhi, Y.~Li, J.~Clune, and K.~Stanley,
  ``Enhanced poet: Open-ended reinforcement learning through unbounded
  invention of learning challenges and their solutions,'' in
  \emph{International Conference on Machine Learning}.\hskip 1em plus 0.5em
  minus 0.4em\relax PMLR, 2020, pp. 9940--9951.

\bibitem{du2022takes}
Y.~Du, P.~Abbeel, and A.~Grover, ``It takes four to tango: Multiagent selfplay
  for automatic curriculum generation,'' \emph{arXiv preprint
  arXiv:2202.10608}, 2022.

\bibitem{dennis2020emergent}
M.~Dennis, N.~Jaques, E.~Vinitsky, A.~Bayen, S.~Russell, A.~Critch, and
  S.~Levine, ``Emergent complexity and zero-shot transfer via unsupervised
  environment design,'' \emph{Advances in neural information processing
  systems}, vol.~33, pp. 13\,049--13\,061, 2020.

\bibitem{parkerholder2022evolving}
J.~Parker-Holder, M.~Jiang, M.~Dennis, M.~Samvelyan, J.~Foerster,
  E.~Grefenstette, and T.~Rocktäschel, ``Evolving curricula with regret-based
  environment design,'' 2022.

\bibitem{fang2021discovering}
K.~Fang, Y.~Zhu, S.~Savarese, and L.~Fei-Fei, ``Discovering generalizable
  skills via automated generation of diverse tasks,'' \emph{arXiv preprint
  arXiv:2106.13935}, 2021.

\bibitem{fang2022active}
K.~Fang, T.~Migimatsu, A.~Mandlekar, L.~Fei-Fei, and J.~Bohg, ``Active task
  randomization: Learning visuomotor skills for sequential manipulation by
  proposing feasible and novel tasks,'' \emph{arXiv preprint arXiv:2211.06134},
  2022.

\bibitem{jiang2021replay}
M.~Jiang, M.~Dennis, J.~Parker-Holder, J.~Foerster, E.~Grefenstette, and
  T.~Rockt{\"a}schel, ``Replay-guided adversarial environment design,''
  \emph{Advances in Neural Information Processing Systems}, vol.~34, pp.
  1884--1897, 2021.

\bibitem{gallouedec2021pandagym}
Q.~Gallou{\'e}dec, N.~Cazin, E.~Dellandr{\'e}a, and L.~Chen, ``{panda-gym:
  Open-Source Goal-Conditioned Environments for Robotic Learning},'' \emph{4th
  Robot Learning Workshop: Self-Supervised and Lifelong Learning at NeurIPS},
  2021.

\bibitem{schulman2017ppo}
J.~Schulman, F.~Wolski, P.~Dhariwal, A.~Radford, and O.~Klimov, ``Proximal
  policy optimization algorithms,'' \emph{arXiv preprint arXiv:1707.06347},
  2017.

\bibitem{haarnoja2018soft}
T.~Haarnoja, A.~Zhou, K.~Hartikainen, G.~Tucker, S.~Ha, J.~Tan, V.~Kumar,
  H.~Zhu, A.~Gupta, P.~Abbeel, \emph{et~al.}, ``Soft actor-critic algorithms
  and applications,'' \emph{arXiv preprint arXiv:1812.05905}, 2018.

\bibitem{stable-baselines3}
\BIBentryALTinterwordspacing
A.~Raffin, A.~Hill, A.~Gleave, A.~Kanervisto, M.~Ernestus, and N.~Dormann,
  ``Stable-baselines3: Reliable reinforcement learning implementations,''
  \emph{Journal of Machine Learning Research}, vol.~22, no. 268, pp. 1--8,
  2021. [Online]. Available: \url{http://jmlr.org/papers/v22/20-1364.html}
\BIBentrySTDinterwordspacing

\end{thebibliography}

\end{document}